# Automatic Summarization of Russian Texts: Comparison of Extractive and Abstractive Methods

**Goloviznina V. S.**
Vyatka State University,
Kirov, Russia
`golovizninavs@gmail.com`

**Kotelnikov E. V.**
Vyatka State University,
Kirov, Russia
`kotelnikov.ev@gmail.com`

**Abstract**

This paper investigates the problem of creating summaries of Russian-language texts based on extractive (TextRank and LexRank) and abstractive (mBART, ruGPT3Small, ruGPT3Large, ruT5-base and ruT5-large) methods. For our experiments, we used the Russian-language corpus of news articles Gazeta and the Russian-language parts of the MLSUM and XL-Sum corpora. We computed ROUGE-N, ROUGE-L, BLEU, METEOR and BERTScore metrics to evaluate the quality of summarization. According to the experimental results, the methods are ranked (from best to worst) as follows: ruT5-large, mBART, ruT5-base, LexRank, ruGPT3Large, TextRank, ruGPT3Small. The study also highlights the salient features of summaries obtained by various methods. In particular, mBART summaries are less abstractive than ruGPT3Large and ruT5-large, and ruGPT3Large summaries are often incomplete and contain errors.

**Keywords:** text summarization; extractive methods; abstractive methods; language models; TextRank; LexRank; mBART; ruGPT3; ruT5

# Автоматическое реферирование русскоязычных текстов: сравнение экстрактивных и абстрактивных методов

**Головизнина В. С.**
Вятский государственный университет,
Киров, Россия
`golovizninavs@gmail.com`

**Котельников Е. В.**
Вятский государственный университет,
Киров, Россия
`kotelnikov.ev@gmail.com`

**Аннотация**

В работе исследуется задача создания рефератов русскоязычных текстов на основе экстрактивных (TextRank и LexRank) и абстрактивных (mBART, ruGPT3Small, ruGPT3Large, ruT5-base и ruT5-large) методов. Для экспериментов использовались русскоязычный корпус новостных статей Gazeta и русскоязычные части корпусов MLSUM и XL-Sum. Для оценки качества реферирования применялись метрики ROUGE-N, ROUGE-L, BLEU, METEOR и BERTScore. По результатам экспериментов методы ранжируются (от лучших к худшим) следующим образом: ruT5-large, mBART, ruT5-base, LexRank, ruGPT3Large, TextRank, ruGPT3Small. Также выделены особенности рефератов, получаемых разными методами. В частности, рефераты mBART оказываются наименее абстрактивными по сравнению с ruGPT3Large и ruT5-large, а рефераты ruGPT3Large часто являются незавершенными и содержат ошибки.

**Ключевые слова:** реферирование текстов; экстрактивные методы; абстрактивные методы; языковые модели; TextRank; LexRank; mBART; ruGPT3; ruT5

## 1 Introduction

Automatic text summarization is the process of creating a summary of the text containing the most important information [5]. There are the following approaches for text summarization – extractive, abstractive and hybrid. With the extractive approach, the summary is formed from the most important sentences of the source text; with the abstractive approach, the content of the summaries is generated and differs from the sentences of the source text. The hybrid approach combines these two approaches. Automatic text summarization methods are used in search engines, to summarize blogs, scientific articles, emails, lawsuits, and medical texts, and to generate headlines for news articles [3].

At present, the choice of the automatic summarization method for the Russian language is not obvious for the following reasons. Firstly, most research is carried out for the English language [27], there are few works for the Russian language [8, 9, 18, 21]. Secondly, a significant part of the works uses only the extractive approach, while abstract methods allow for a shorter and human-like presentation that differs from the sentences of the original text [3]. Thirdly, a number of new Russian-language neural network models have recently appeared, such as ruGPT-3 and ruT5, which have not been sufficiently studied in the summarization task.

Thus, the task of conducting a comparative analysis of extractive and abstractive methods of summarization on the material of the Russian language, including modern language models, is relevant.

The contribution of this work is as follows:
- for the first time, there has been carried a simultaneous comparison of extractive (TextRank and LexRank) and abstractive (mBART, ruGPT-3 and ruT5) summarization methods using three corpora of news articles: Gazeta [8], MLSUM [27] and XL-Sum [9];
- the methods under investigation have been ranked based on the ROUGE-N, ROUGE-L, BLEU, METEOR and BERTScore quality metrics;
- the salient features of summaries obtained by different methods have been revealed.

The paper is structured as follows. The second section provides an overview of previous work on Russian texts summarization. The third section is devoted to text corpora, models and methods used for text summarization. In the fourth section the experimental results are presented and discussed. The fifth section provides conclusions and suggests directions for further research.

## 2 Previous work

Language models based on the Transformer architecture [28] have become a key technology for solving natural language processing problems, including automatic text summarization [15]. Such models as mBART [16], ruGPT3 [25], and mT5 [29] have been used for summarizing Russian-language texts.

Gusev [8] fine-tuned the multilingual mBART model for text summarization on the Russian-language Gazeta dataset. The model showed the best results among abstractive models in ROUGE and BLEU metrics. In addition to mBART, Gusev used Pointer-generator, CopyNet models and extractive methods TextRank, LexRank and LSA.

Nikolich et al. [18] used the ruGPT3Small model, fine-tuned on Gazeta corpus, for text summarization in Russian. ruGPT3Small outperformed mBART [8] only in BERTScore.

Hasan et al. [9] fine-tuned the mT5 model for summarization in 44 languages, including Russian, using the XL-Sum corpus. The results of mT5 are close to the current level of summarization in English [30]. ROUGE-2 scores for other languages are comparable to results in English.

Polyakova and Pogoreltsev [21] proposed a new method of extractive summarization that reduces the problem to selecting the most probable sequence of sentences. The method outperforms the SummaRuNNer and mBART models in ROUGE-1 and ROUGE-L on the Gazeta dataset.

In our paper, in contrast to [8], besides mBART, we fine-tuned ruGPT3Small, ruGPT3Large [25], ruT5-base and ruT5-large [26] models. In contrast to [18], we used not only ruGPT3Small, but also the ruGPT3Large model. In contrast to [9], instead of multilingual mT5 model, we applied the Russian-language ruT5-large model. In contrast to [21], abstractive models are studied. Besides, it is the first time that all these methods and models are simultaneously analyzed using the three corpora: Gazeta, MLSUM, and XL-Sum.

## 3 Materials and Methods

### 3.1 Text Corpora

Corpora for text summarization are sets of texts and summaries to them. Our study uses the Russian-language corpus of news articles Gazeta and the Russian-language parts of the MLSUM and XL-Sum corpora. The Gazeta corpus consists of 63,435 articles from the news source Gazeta.ru[1] [8]. The MLSUM corpus contains 1,259,096 articles in five languages (German, Spanish, French, Russian, Turkish), of which 27,063 articles are in Russian from "Moskovsky Komsomolets"[2] [27]. XL-Sum consists of BBC[3] news articles in 45 languages and contains about 1,350,000 articles, of which 77,803 are in Russian [9]. Characteristics of the corpora are shown in Table 1.

| Corpus (source) | Dataset | Size | Data | Length in tokens | | |
|---|---|---|---|---|---|---|
| | | | | min | max | mean |
| Gazeta (Gazeta.ru) | train | 52,400 (82.6%) | text | 28 | 1,500 | 766.5 |
| | | | summary | 15 | 85 | 48.8 |
| | validation | 5,265 (8.3%) | text | 191 | 1,500 | 772.4 |
| | | | summary | 18 | 85 | 54.5 |
| | test | 5,770 (9.1%) | text | 357 | 1,498 | 750.3 |
| | | | summary | 18 | 85 | 53.2 |
| MLSUM[4] ("Moskovsky Komsomolets") | train | 25,556 (94.4%) | text | 55 | 11,689 | 949.9 |
| | | | summary | 10 | 65 | 14.7 |
| | validation | 750 (2.8%) | text | 118 | 5,842 | 1,156.7 |
| | | | summary | 10 | 30 | 13.4 |
| | test | 757 (2.8%) | text | 69 | 26,794 | 1,214.4 |
| | | | summary | 10 | 35 | 13.4 |
| XL-Sum[5] (BBC News) | train | 62,243 (80%) | text | 19 | 22,274 | 682.1 |
| | | | summary | 1 | 246 | 29.4 |
| | validation | 7,780 (10%) | text | 62 | 1,583 | 556.9 |
| | | | summary | 8 | 60 | 27.9 |
| | test | 7,780 (10%) | text | 54 | 1,745 | 555.8 |
| | | | summary | 8 | 60 | 27.9 |

Table 1: Corpora statistics.
The length in tokens is specified for the *razdel*[6] tokenizer.

### 3.2 Extractive methods

For extractive summarization, we used TextRank method from the *summa* library[7] [2] and LexRank from the *lexrank* library[8].

---

[1] https://www.gazeta.ru.
[2] https://www.mk.ru/news.
[3] https://www.bbc.com.
[4] MLSUM corpus has a very large max length of the texts (26,794 tokens) but it contains only 54 texts (0.2%) with a length of more than 5,000 tokens.
[5] The training part of XL-Sum has a very large max length of the texts (22,274 tokens) and very salient min and max lengths for train summaries (min=1 and max=246). But it contains only 96 texts (0.15%) with a length exceeding 5,000 tokens and only 5 summaries with a length of less than 3 tokens, and 23 summaries with a length of more than 100 tokens.
[6] https://natasha.github.io/razdel.
[7] https://pypi.org/project/summa.
[8] https://github.com/crabcamp/lexrank.

TextRank [17] is a method used for keyword extraction and extractive summarization. In the method, the text is divided into sentences, between which the similarity is calculated, and the PageRank algorithm [19] is used to obtain sentence scores. The sentences with the highest scores are included in the summary. A measure of the sentence similarity is the number of common words in these sentences.

In the LexRank method [4], the similarity measure of sentences is the cosine similarity of the TF-IDF vectors of these sentences. The method uses the following idea: if a sentence is similar to other sentences, then it is the central sentence of this text, that is, it contains the necessary and sufficient information about the entire text.

### 3.3 Abstractive methods

For abstractive summarization, we applied mBART, ruGPT3Small, ruGPT3Large, ruT5-base and ruT5-large models.

The BART (Bidirectional and Auto-Regressive Transformer) model is based on the Transformer architecture and includes a bidirectional encoder (like BERT) and an autoregressive decoder (like GPT) [13]. Two model versions are available: $BART_{BASE}$ and $BART_{LARGE}$. The multilingual version mBART was trained on the Common Crawl corpus[9] for 25 languages. We used the multilingual $mBART_{LARGE}$, fine-tuned for text summarization on the Gazeta dataset [8].

Models of the GPT (Generative Pre-trained Transformer) family consist of a Transformer decoder with a different number of layers [22]. The family includes three main models: GPT [22], GPT-2 [23], and GPT-3 [1]. The ruGPT-3 model is a Russian-language model from Sber based on GPT-2 [25]. The model was trained on 80 billion tokens. There are five versions of different sizes: ruGPT3Small, ruGPT3Medium, ruGPT2Large, ruGPT3Large, and ruGPT3XL. In our experiments, we used the ruGPT3Small and ruGPT3Large model.

The T5 (Text-to-Text Transfer Transformer) model was trained on 24 tasks for the English language [24]. The multilingual version mT5 was trained for 101 languages, but on one task – text filling. The ruT5 model is a Russian-language T5 model from Sber, available in two versions: ruT5-base and ruT5-large [26]. The model was trained on the same corpus as ruGPT-3. We used both versions: ruT5-base and ruT5-large.

## 4 Experiments

### 4.1 Experimental Setup

The TextRank method was applied with *compression ratio* = 0.2 (default value). For the LexRank method, the length of the summary was limited to three sentences (*summary_size* = 3). The rest of the methods parameters assumed default values.

During the experiments with each corpus, $mBART_{LARGE}$, ruGPT3Small, ruGPT3Large, ruT5-base and ruT5-large models were fine-tuned on the training part of the given corpus. The validation part of the corpus was used to select the number of training epochs. For the $mBART_{LARGE}$, ruT5-base and ruT5-large models, the length of the input text was 1,024 tokens, the length of the output data (the length of the generated summary) was limited by the length of the reference summary. The desired size of the summary is often a requirement in real-world problems. Given the availability of reference summary in our experiments, it is logical to use their size as a limitation.

For the ruGPT3Small and ruGPT3Large models, the length of the output data was regulated in the same way, the length of the input data was 2,048 tokens. When fine-tuning, the input of the ruGPT3Small and ruGPT3Large models was given sequences of the form: "Text:*text*[SEP]Summary:*summary*", where *text* is the input text, *summary* is the reference summary for this text. When testing, the model generated a summary for the following input: "Text:*text*[SEP]Summary:".

We also tested Lead-3 – it is a strong baseline, where summary is the first three sentences of every text.

We used five automatic metrics: ROUGE-N [14], ROUGE-L [14], BLEU [20], METEOR [12], and BERTScore [30] to evaluate the results. To calculate the ROUGE-N and ROUGE-L metrics, we applied the *rouge* library[10], for BLEU and METEOR – the *NLTK* library[11], *Snowball Stemmer*[12], and the *wiki_ru_wordnet* semantic network[13]. BERTScore uses embeddings from BERT and matches words in generated summaries and reference summaries by cosine similarity. We calculated BERTScore using the *bert-score* library[14] and the Russian-language RuBERT model [10].

---

[9] https://commoncrawl.org.

### 4.2 Results and Discussion

Table 2 shows the results of experiments for the three corpora, as well as the average values.
According to the results of experiments (see Table 2), models and methods can be ranked as follows:
1. ruT5-large,
2. mBART,
3. ruT5-base,
4. LexRank,
5. ruGPT3Large,
6. TextRank,
7. ruGPT3Small.

The ruT5-large and mBART models showed the best results, but mBART tends to repeat parts of the source text sentences. Figure 1 shows the average proportion of novel n-grams for three corpora in the summaries of abstractive models. A novel n-gram is an n-gram of the summary that is not contained in the source text. The proportion of novel n-grams is the number of novel n-grams divided by all n-grams of the summary. mBART summaries have the smallest proportion of novel n-grams. The proportion of novel n-gram summaries of ruT5-base and ruT5-large is greater than the proportion of n-grams of mBART summaries, but never more than the proportion of novel n-grams of reference summaries, which is surpassed by ruGPT3Small and ruGPT3Large. The ruGPT3Small and ruGPT3Large summaries contain the largest proportion of novel n-grams, but there are often errors – mismatches between the summary and the source text. Despite the large proportion of novel n-grams, ruT5-large summaries have significantly fewer errors than ruGPT-3Large summaries.

| Corpus | Method | ROUGE-1 | ROUGE-2 | ROUGE-L | BLEU | METEOR | BERTScore |
|---|---|---|---|---|---|---|---|
| Gazeta | Lead-3 | 31.02 | 13.44 | 27.69 | 10.80 | **34.44** | 56.49 |
| | TextRank | 21.44 | 6.27 | 18.56 | 3.92 | 26.31 | 49.90 |
| | LexRank | 23.93 | 8.00 | 20.96 | 5.64 | 28.17 | 51.49 |
| | mBART | 31.55 | 13.54 | 28.22 | **11.19** | 34.09 | 56.56 |
| | ruT5-base | 30.45 | 12.63 | 27.41 | 9.54 | 28.69 | 56.35 |
| | ruT5-large | **32.45** | **13.97** | **29.24** | 10.88 | 31.21 | **57.73** |
| | ruGPT3Small | 18.84 | 4.06 | 16.68 | 3.13 | 18.70 | 44.06 |
| | ruGPT3Large | 23.45 | 6.45 | 20.73 | 4.93 | 23.77 | 47.76 |
| MLSUM | Lead-3 | 9.42 | 1.55 | 8.47 | 0.86 | **12.98** | 32.15 |
| | TextRank | 4.76 | 0.55 | 4.39 | 0.13 | 7.51 | 29.22 |
| | LexRank | 10.22 | 1.42 | 7.36 | 0.90 | 11.28 | 31.83 |
| | mBART | 11.48 | 1.95 | 10.26 | 1.49 | 10.52 | 37.89 |
| | ruT5-base | 12.35 | 1.86 | 11.22 | 1.58 | 9.68 | 38.67 |
| | ruT5-large | **14.06** | **2.86** | **12.69** | **2.81** | 11.84 | **39.92** |
| | ruGPT3Small | 9.14 | 0.60 | 8.13 | 0.40 | 6.66 | 34.27 |
| | ruGPT3Large | 9.36 | 0.99 | 8.17 | 0.73 | 7.44 | 35.00 |
| XL-Sum | Lead-3 | 16.14 | 3.38 | 13.57 | 1.63 | 22.70 | 46.29 |
| | TextRank | 14.04 | 3.14 | 11.81 | 1.05 | 21.45 | 45.80 |
| | LexRank | 16.22 | 3.16 | 12.69 | 2.14 | 17.20 | 43.83 |
| | mBART | 26.47 | 10.95 | 22.67 | 7.51 | 27.16 | 54.24 |
| | ruT5-base | 26.52 | 10.67 | 22.79 | 6.58 | 25.35 | 52.89 |
| | ruT5-large | **28.42** | **11.98** | **24.41** | 7.93 | **28.31** | **56.06** |

---
[10] https://pypi.org/project/rouge.
[11] https://www.nltk.org.
[12] https://snowballstem.org.
[13] https://wiki-ru-wordnet.readthedocs.io/en/latest.
[14] https://github.com/Tiiiger/bert_score.

| Corpus | Method | ROUGE-1 | ROUGE-2 | ROUGE-L | BLEU | METEOR | BERTScore |
|---|---|---|---|---|---|---|---|
| XL-Sum | ruGPT3Small | 16.19 | 3.28 | 13.68 | 2.25 | 15.94 | 40.12 |
| | ruGPT3Large | 19.37 | 5.17 | 16.48 | 3.74 | 19.63 | 42.74 |
| Average | Lead-3 | 18.86 | 6.12 | 16.58 | 4.43 | 23.37 | 44.98 |
| | TextRank | 13.41 | 3.32 | 11.59 | 1.70 | 18.42 | 41.64 |
| | LexRank | 16.79 | 4.19 | 13.67 | 2.89 | 18.88 | 42.38 |
| | mBART | 23.17 | 8.81 | 20.38 | 6.73 | **23.92** | 49.56 |
| | ruT5-base | 23.11 | 8.39 | 20.47 | 5.90 | 21.24 | 49.30 |
| | ruT5-large | **24.98** | **9.60** | **22.11** | **7.21** | 23.79 | **51.24** |
| | ruGPT3Small | 14.72 | 2.65 | 12.83 | 1.93 | 13.77 | 39.48 |
| | ruGPT3Large | 17.39 | 4.20 | 15.13 | 3.13 | 16.95 | 41.83 |

Table 2: Automatic summarization scores on Gazeta, MLSUM and XL-Sum corpora

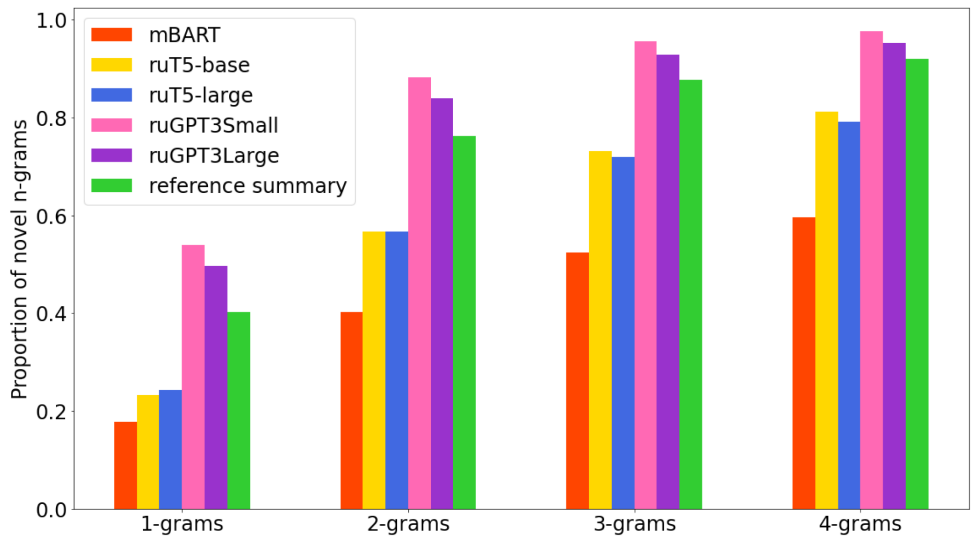

Figure 1: Average proportion of novel n-grams in summaries of abstractive models

The conclusions on the proportion of novel n-grams in summaries are confirmed by the extraction score [6] shown in Table 3. Extraction score is the sum of the normalized lengths of all long non-overlapping common sequences between a text and a summary, which is in the range from 0 to 1. This metric is inversely proportional to the degree of abstractiveness of the constructed summary. The lowest level of abstractiveness is shown by the mBART model, the highest – by ruGPT3Small.

| Corpus | Characteristic | Reference summary | mBART | ruGPT3 Small | ruGPT3 Large | ruT5-base | ruT5-large |
|---|---|---|---|---|---|---|---|
| Gazeta | Average length of summaries in tokens | 53.2 | 59.8 | 54.0 | 54.8 | 42.3 | 44.5 |
| | Extraction score | 0.06 | 0.39 | 0.03 | 0.05 | 0.24 | 0.26 |
| MLSUM | Average length of summaries in tokens | 13.4 | 18.8 | 13.2 | 13.5 | 12.4 | 15.0 |
| | Extraction score | 0.09 | 0.39 | 0.05 | 0.06 | 0.16 | 0.17 |
| XL-Sum | Average length of summaries in tokens | 27.9 | 22.2 | 29.0 | 28.9 | 19.8 | 21.5 |
| | Extraction score | 0.04 | 0.10 | 0.03 | 0.04 | 0.08 | 0.07 |
| Average | Average length of summaries in tokens | 31.5 | 33.6 | 32.1 | 32.4 | 24.8 | 27.0 |
| | Extraction score | 0.06 | 0.29 | 0.04 | 0.05 | 0.16 | 0.17 |

Table 3: Average summaries lengths (in tokens) and extraction score. The length in tokens is specified for the *razdel* tokenizer. Smaller values of extraction score correspond to a greater degree of abstractiveness of summaries

Another problem with abstractive methods is the incompleteness of the generated summaries. The ruGPT3Small and ruGPT3Large generate summaries that are closest in length to the reference ones (Table 3), but often does not complete them, while ruT5-base and ruT5-large, as a rule, complete sentences. Table 4 shows the proportion of summaries that do not end in end-of-sentence punctuation marks: ".", "!", "?". For MLSUM, this value was not calculated, since the reference summaries from which the models were trained do not have punctuation marks at the end of the last sentence.

| Corpus | mBART | ruGPT3Small | ruGPT3Large | ruT5-base | ruT5-large |
|--------|-------|-------------|-------------|-----------|------------|
| Gazeta | 0.10  | 0.86        | 0.96        | 0.09      | 0.14       |
| XL-Sum | 0.42  | 0.90        | 0.95        | 0.19      | 0.02       |

Table 4: The proportion of summaries that do not end in one of the punctuation marks ".", "!", "?"

With regard to extractive methods, LexRank performed better than TextRank (see Table 2). Figure 2 for extractive methods shows the proportion of extracted sentences according to their position in the source text. TextRank selects sentences from the text more evenly, LexRank tends to select sentences from the beginning of the text. Both methods include the first sentence of the text in summaries more often than others. This is due to the structure of the news article – the main information is contained at the beginning of the text, and then the clarifying facts are indicated.

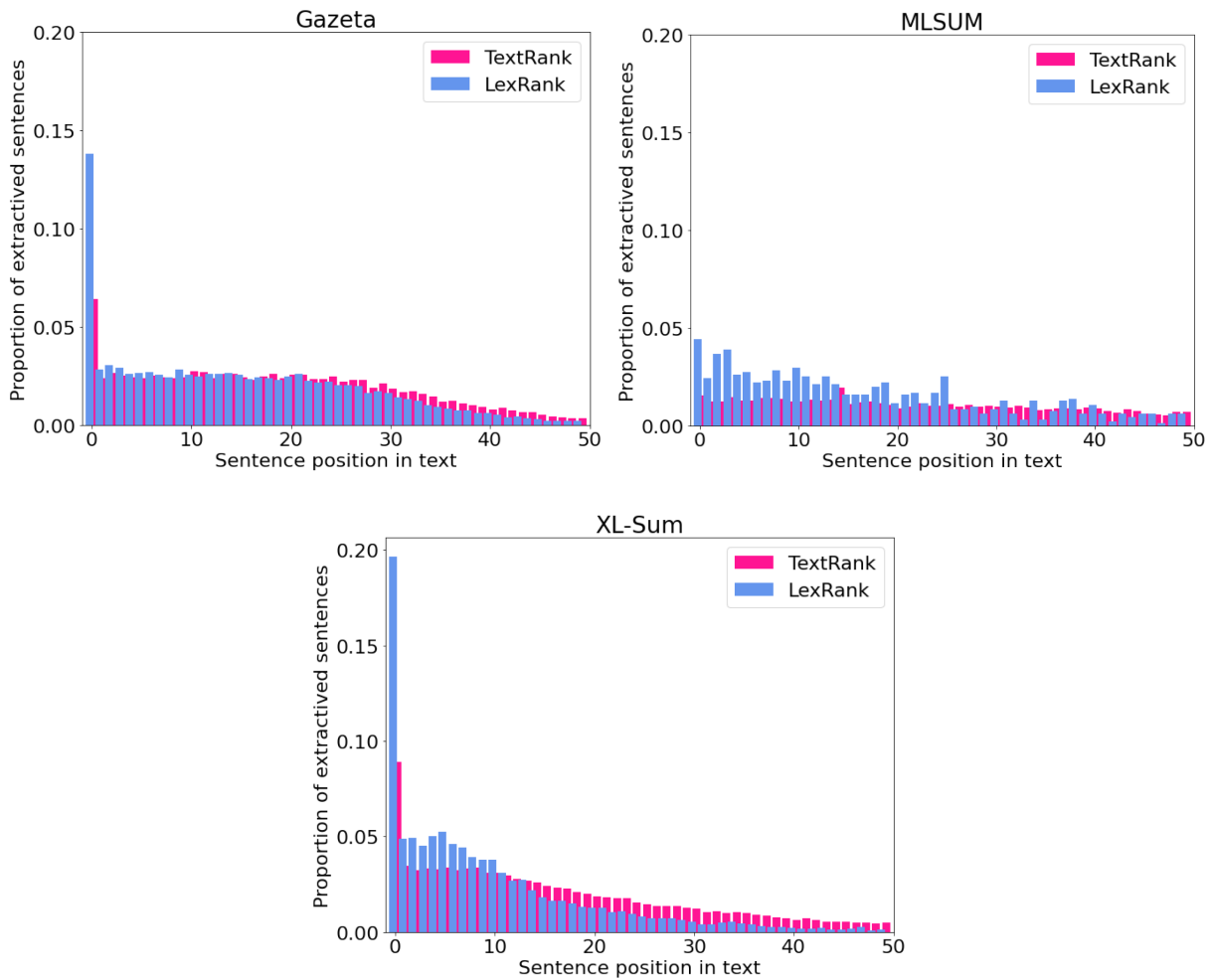

Figure 2: The dependence of the proportion of extracted sentences according to their position in the source text

The comparison between LexRank and ruGPT3Large is ambiguous (see Table 2). For the Gazeta corpus, the LexRank method outperforms ruGPT3Large in all metrics, for the MLSUM corpus it outperforms in 4 out of 6 metrics, for the XL-SUM corpus it is inferior in all metrics, except for BERTScore. On average, LexRank is ahead of ruGPT3Large in terms of METEOR and BERTScore, and for the ROUGE-2 metric the results differ by 0.01. However, we have decided to rank LexRank higher than ruGPT3Large due to the large number of factual errors of the latter, which cannot be in the extractive method. For the same reason, we put TextRank higher than ruGPT3Small.

Figures 3 (Russian version) and 4 (English version) show examples of summaries created by all eight methods. In the reference summary for the text from the Gazeta, two ideas stand out – a description of the new method and its criticism. It is only in the summary obtained by LexRank that there is an attempt to retain both ideas. The remaining methods pay attention to the first idea, while most of the mBART summary repeats the source text, the ruGPT3Large summary is not completed and contains errors.

From Table 2 it can be seen that for MLSUM the metrics are low compared to the other two corpora. This can be explained by the fact that MLSUM is different from the other two datasets: a) MLSUM is 2.3 times smaller than Gazeta and 2.9 times smaller than XLSUM; b) the average length of a summary in MLSUM is much shorter, it is 13.8 tokens, while the average length in Gazeta is 52.2 tokens, in XLSUM it is 28.4 tokens.

We tried to evaluate the adequacy of the automatic metrics. We used four criteria proposed by Fabbri et al. [5]: coherence, consistency, fluency and relevance. We also added included two additional criteria: no excess information (absence of redundancy in the summary) and abstractiveness (absence of sentences from the source text). We randomly selected five texts for each of the corpora and evaluated summaries generated using all eight methods, that is, a total of 5*3*8=120 examples. During the evaluation, text, title, reference summary and generated summary were available to annotators. The annotator for each generated summary gave a score on a scale from 1 to 5 for each of the 6 criteria. We computed Kendall's tau rank correlations between average manual scores and all the automatic metrics. The correlation values turned out quite high – from 0.7143 (ROUGE-3) to 0.9286 (BERTScore). This confirms the adequacy of using automatic evaluation metrics.

| Text / summary | Gazeta_594[15] |
|---|---|
| text[16] | Для поимки преступника хватит и пары волосков, заявляют специалисты из Национального института стандартов и технологий в США. Разработанный ими метод идентификации способен выделить специфические для конкретного человека белки из пряди волос длиной всего сантиметр. Для других аналогичных подходов требуется в восемь раз большая длина. О новом способе исследователи рассказали в статье в журнале Journal of Forensic Sciences. … Поскольку последовательности аминокислот в белках различаются от человека к человеку, такой подход позволит с высокой степенью точности идентифицировать людей и без выделения ДНК… |
| reference summary | Установить личность преступника можно всего по паре волосков, утверждают американские ученые — это можно сделать даже не выделяя ДНК, а лишь по ряду белковых последовательностей. Однако метод занимает слишком много времени, поэтому неизвестно, найдет ли он применение в криминалистике. |
| Lead-3 | Для поимки преступника хватит и пары волосков, заявляют специалисты из Национального института стандартов и технологий в США. Разработанный ими метод идентификации способен выделить специфические для конкретного человека белки из пряди волос длиной всего сантиметр. Для других аналогичных подходов требуется в восемь раз большая длина. |
| TextRank | Однако большинство методов исследования белков в волосах подразумевают многократное измельчение и нагрев волос, в результате чего большая часть белков разрушается, а оставшихся может не хватить для получения необходимых данных. Точнее такую возможность просчитать пока сложно — в этом исследовании использовались образцы волос из биобанка и авторы не ставили перед собой цель найти их обладателя. В связи с этим возникает ряд вопросов — как краска для волос и другие косметические продукты влияют на эти белки? |
| LexRank | Однако большинство методов исследования белков в волосах подразумевают многократное измельчение и нагрев волос, в результате чего большая часть белков разрушается, а оставшихся может не хватить для получения необходимых данных. Разработанный ими метод идентификации способен выделить специфические для конкретного человека белки из пряди волос длиной всего сантиметр. Метод идентификации личности по полученным из волос GVP ранее не применялся в судебной практике, подчеркивает судмедэксперт Монти Миллер. |
| mBART | В США разработали новый способ идентификации личности по волосам — он позволяет выделить специфические для конкретного человека белки из пряди волос длиной всего сантиметр. Такой подход позволит с высокой степенью точности идентифицировать людей и без выделения ДНК. |
| ruGPT3Small | Исследование позволяет выявить особенности внешности преступника с помощью двух методов, направленных на выявление возможных причин, вызвавших задержку ДНК в ДНК. Миллионы людей в мире смогут идентифицировать себя по ДНК и другим типам волос |
| ruGPT3Large | Проверить человека по волосатой персоне в настоящее время можно достаточно просто. Теперь ученые разработали новый метод, позволяющий выявить человека по его волосяным фолликулам. Ранее для установления личности по образцам волос нужно было идентифицировать ДНК из кожи, с которой был |
| ruT5-base | Ученые выяснили, как определить личность человека по волосам. Для этого им достаточно выделить несколько волосков, чтобы поимка преступника была более точной. |
| ruT5-large | Американские ученые разработали новый способ идентификации личности по волосам. Для этого достаточно выделить белки из пряди волос длиной всего сантиметр. |

Figure 3: Example of summaries received for the text from Gazeta (Russian version).
Yellow – repeated text. Blue – the second idea of the reference summary, which is not reflected in any summary of abstractive models. Red – the summary is not completed. Green – the most important and correct information in the summary. Red – errors. Orange – defect (pronouns in summaries of extractive methods)

---

[15] https://www.gazeta.ru/science/2019/11/24_a_12829868.shtml.
[16] The text is not given in full: its main sentences are included in the example.

| Text / summary | Gazeta_594[117] |
|---|---|
| text[18] | A couple of hairs is enough to catch a criminal, say experts from the National Institute of Standards and Technology in the United States. The identification method they developed is capable of isolating human-specific proteins from a strand of hair just a centimeter long. Other similar approaches require eight times the length. The researchers described the new method in an article in the Journal of Forensic Sciences. … Since the sequences of amino acids in proteins differ from person to person, this approach will allow people to be identified with a high degree of accuracy without DNA isolation… |
| reference summary | It is possible to establish the identity of a criminal with just a couple of hairs, according to American scientists — this can be done even without isolating DNA, but only by a number of protein sequences. However, the method takes too much time, so it is not known whether it will find application in forensic science. |
| Lead-3 | A couple of hairs is enough to catch a criminal, say experts from the National Institute of Standards and Technology in the United States. The identification method they developed is capable of isolating human-specific proteins from a strand of hair just a centimeter long. Other similar approaches require eight times the length. |
| TextRank | However, most methods for studying hair proteins involve repeated grinding and heating of the hair, as a result of which most of the proteins are destroyed, and the remaining ones may not be enough to obtain the necessary data. It is still difficult to calculate this possibility more precisely - in this study, hair samples from the biobank were used and the authors did not set themselves the goal of finding their owner. In this regard, a number of questions arise - how do hair dye and other cosmetic products affect these proteins? |
| LexRank | However, most methods for studying hair proteins involve repeated grinding and heating of the hair, as a result of which most of the proteins are destroyed, and the remaining ones may not be enough to obtain the necessary data. The identification method they developed is capable of isolating human-specific proteins from a strand of hair just a centimeter long. The method of identifying a person using GVP obtained from hair has not previously been used in judicial practice, emphasizes forensic scientist Monty Miller. |
| mBART | In the United States, a new method of identifying a person by hair has been developed — it allows to isolate human-specific proteins from a strand of hair just a centimeter long. This approach will allow people to be identified with a high degree of accuracy without DNA isolation. |
| ruGPT3Small | The study allows you to identify the features of the criminal's appearance using two methods aimed at identifying possible causes that caused the DNA to be retained in DNA. Millions of people in the world will be able to identify themselves by DNA and other types of hair |
| ruGPT3Large | Checking a person by a hairy person is currently quite simple. Now scientists have developed a new method to identify a person by their hair follicles. Previously, in order to establish an identity from hair samples, it was necessary to identify DNA from the skin with which |
| ruT5-base | Scientists have figured out how to determine the personality of a person by hair. To do this, it is enough for them to select a few hairs so that the capture of the criminal is more accurate. |
| ruT5-large | American scientists have developed a new method of identifying a person by hair. To do this, it is enough to isolate proteins from a strand of hair just a centimeter long. |

Figure 4: Example of summaries received for the text from Gazeta (English version).
Yellow – repeated text. Blue – the second idea of the reference summary, which is not reflected in any summary of abstractive models. Red – the summary is not completed. Green – the most important and correct information in the summary. Red – errors. Orange – defect (pronouns in summaries of extractive methods)

---

[17] https://www.gazeta.ru/science/2019/11/24_a_12829868.shtml.
[18] The text is not given in full: its main sentences are included in the example.

## 4.3 Comparison with other works

Comparison of our results with the results of [8, 9, 18] is difficult. Hasan et al. [9] give the values of their own modified ROUGE metric, which considers the language – multilingual rouge scoring[19], while we calculate the standard ROUGE metric [14]. Also, as a metric that takes into account the language, we use METEOR with Russian-language *Snowball Stemmer* and the *wiki_ru_wordnet* semantic network. Gusev [8] uses a different METEOR library. In addition, in [8] the input length was limited to 600 tokens, in our work – to 1024. Nikolich et al. [18] calculate BERTScore using the multilingual BERT model [7], we use RuBERT [10]. The values of the parameters for language models are different in our work and, for example, in [18].

In this regard, we show our results along with the results of [8, 18, 21] only on ROUGE and BLEU metrics (Table 5). The Gazeta is the only corpus, which these works investigate. To emphasize the difficulty of direct comparison, we did not highlight the best results in Table 5.

| Model | ROUGE-1 | ROUGE-2 | ROUGE-L | BLEU |
|---|---|---|---|---|
| ruT5-large (our work, Table 2) | 32.45 | 13.97 | 29.24 | 10.88 |
| mBART [8] | 32.11 | 14.2 | 27.9 | 12.4[20] |
| ruGPT3Large (our work, Table 2) | 23.45 | 6.45 | 20.73 | 4.93 |
| ruGPT3Small (our work, Table 2) | 18.84 | 4.06 | 16.68 | 3.13 |
| ruGPT3Small [18] | 11.4 | 1.4 | 10.0 | 23.1[21] |
| [21] | 35.6 | 14.2 | 32.4 | – |

Table 5: The comparison of our results with other works for the Gazeta corpus

## 5 Conclusion

In the study, we compared several models and methods within the framework of abstractive and extractive approaches on the corpora of news articles Gazeta, MLSUM and XL-Sum.

Based on the experimental results, we ranked the methods (from best to worst) as follows: ruT5-large, mBART, ruT5-base, LexRank, ruGPT3Large, TextRank, ruGPT3Small.

During the analysis of summaries obtained by different methods, we identified several features:
- mBART has the lowest level of abstractiveness compared to ruGPT3Large and ruT5-large;
- ruGPT3Small and ruGPT3Large generate summaries that are closest in length to the reference ones, but often does not complete them and makes errors;
- ruT5-base and ruT5-large summaries are also close to the reference length, rather abstract and contain fewer errors than summaries of ruGPT3Small and ruGPT3Large;
- TextRank more evenly selects sentences from the text;
- LexRank tends to select sentences from the beginning of the text.

In further research, we intend to compare the considered methods on the Russian-language part of the WikiLingva corpus [11], formed on WikiHow articles, which differ in their structure from news articles.

## Acknowledgements

This work was supported by Russian Science Foundation, project № 22-21-00885,
https://rscf.ru/en/project/22-21-00885.

---

[19] https://github.com/csebuetnlp/xl-sum/tree/master/multilingual_rouge_scoring.
[20] Gusev [8] made a mistake in calculating BLEU. Updated scores: https://arxiv.org/pdf/2006.11063.pdf
[21] In [18], the BLEU value is compared with the erroneous results of [8], it is probably also incorrect.